# An efficient genetic algorithm for large-scale transmit power control of dense industrial wireless networks

Xu Gong*, David Plets, Emmeric Tanghe, Toon De Pessemier, Luc Martens, Wout Joseph

*Department of Information Technology, Ghent University/imec, Technologiepark 15, 9052 Ghent, Belgium*

* E-mail: xu.gong@ugent.be. Tel.: +32 4 88 69 68 01. Fax.: +32 9 33 14899.

**Abstract**

The industrial wireless local area network (IWLAN) is increasingly dense, not only due to the penetration of wireless applications into factories and warehouses, but also because of the rising need of redundancy for robust wireless coverage. Instead of powering on all the nodes with the maximal transmit power, it becomes an unavoidable challenge to control the transmit power of all wireless nodes on a large scale, in order to reduce interference and adapt coverage to the latest shadowing effects in the environment. Therefore, this paper proposes an efficient genetic algorithm (GA) to solve this transmit power control (TPC) problem for dense IWLANs, named GATPC. Effective population initialization, crossover and mutation, parallel computing as well as dedicated speedup measures are introduced to tailor GATPC for the large-scale optimization that is intrinsically involved in this problem. In contrast to most coverage-related optimization algorithms which cannot deal with the prevalent shadowing effects in harsh industrial indoor environments, an empirical one-slope path loss model considering three-dimensional obstacle shadowing effects is used in GATPC, in order to enable accurate yet simple coverage prediction. Experimental validation and numerical experiments in real industrial cases show the promising performance of GATPC in terms of scalability to a hyper-large scale, up to 37-times speedup in resolution runtime, and solution quality to achieve adaptive coverage and to minimize interference.

# 1. Introduction

The dominant wireless local area network (WLAN) technology IEEE802.11 or WiFi is penetrating into factories to promote factories of the future (FoF) [1]. Compared to cabled technologies for interconnection of machines or devices, wireless technologies are superior regarding mobility, flexibility and cheap installation and maintenance. Compared to other wireless technologies, IEEE802.11 has the advantages of low cost, high data rate and considerable coverage distance.

An industrial WLAN (IWLAN) is emerging as a basic infrastructure for manufacturing operations. For instance, production cell controllers can connect to other intelligent devices such as robot arms via an IWLAN on the shop floor [2], in order to realize agile production. The other industrial operations that are increasingly relying on IWLANs are illustrated as intra-factory transportation by automated guided vehicles (AGVs), video monitoring, process monitoring, etc.

However, a typical industrial indoor environment is harsh in terms of radio propagation. Either a shop floor or a warehouse is dominant by various metal facilities, such as production machines/lines, storage racks, steel bars, metal plates, pipes, AGVs, cranes and forklifts. These metals easily shadow radio propagation [1, 3]. Consequently, this creates coverage hole for a WLAN that is already deployed. Moreover, an industrial indoor layout may occasionally be altered with the prevalence of flexible manufacturing [4]. This makes it increasingly difficult to maintain the expected wireless coverage in the presence of these shadowing effects. Therefore, it is of strategic importance to conceive an effective method to tackle these shadowing effects for robust wireless connection of personnel, machines, materials and products in FoF.

Furthermore, an IWLAN is denser compared to a public WLAN. This is not only due to the large size of an industrial indoor environment, but also driven by the increasing industrial need for redundant coverage to ensure high network availability [5]. While large-scale optimization is increasingly desired [6, 7], most research on coverage optimization problems neglects the scalability of an optimization algorithm [8-12]. In addition, these studies tend to focus on optimization problems of wireless sensor networks (WSNs), although dense WLANs are showing up their application significance [13, 14].

Several powering-on/off solutions for dense WLANs have been introduced in literature, to enable energy-efficient dense WLANs. A concept of resource on demand (RoD) was proposed in [15], where redundant APs are powered off when they are detected to remain idle according to the volume and location of user demand. A more general model was proposed in [16] to further demonstrate the effectiveness of RoD. Energy savings up to 87% were proven to be achieved during low-traffic periods, with hardly any sacrifice in QoS. However, besides the powering-on/off mechanism, the idea of transmit power control (TPC) was only highlighted and not investigated in these studies [15, 16]. In addition, authors in [17] mentioned that there is little commercial off-the-shelf (COTS) hardware to support the powering-on/off mechanism. This causes another challenge for the management of dense WLANs.

Cell breathing by TPC is a well-known concept in cellular networks [18, 19]. For instance, authors in [19] investigated a problem of minimizing total WCDMA pilot power subject to a coverage constraint. A WCDMA cell shrinks or expands according to the varying coverage rate, following the trade-off between power consumption and coverage. Nevertheless, many critical aspects are missing which prevent this work to be analogously applied to a dense IWLAN: a proper path loss (PL) model that considers obstacle shadowing, a large-scale problem size, and powering-on/off mechanism.

Conversely, cell breathing of WLANs is found in little literature. Power management algorithms were proposed in [20] to control the coverage of APs. However, without using any PL model, the authors assumed that the received power of a client is proportional to the transmission power of the connected AP. Analogously, a lack of proper PL modeling can be observed in [21]. A TPC scheme was proposed, but only a linear approximation was assumed between the AP transmission power and RSSI (received signal strength indictor) of a client.

In WSN coverage related optimization problems, the classical Boolean disk model is widely used to calculate coverage [8, 22]. It is simple by only considering a circular area within which all GPs are coverage. But its application to the IWALN coverage related optimization problems could drastically simplify the problem description and degrade the optimal solution's quality, since it ignores the obstacle shadowing effects and cannot calculate the exact received RF power of a GP in the target environment. This RF power is further linked to interference estimation, which is an indispensable concern for dense WLANs [23]. On the other hand, it is costly and time consuming to undertake a complete site survey, in order to capture the actual coverage and interference. As highlighted in [24], a combination of site survey and planning algorithm design is a good method to reduce the required measurements without compromising much the coverage prediction.

This paper proposes a large-scale TPC problem for dense IWLANs. The contributions are fourfold. (1) An empirical one-slope PL model is introduced for precise yet simple coverage calculation, including the obstacle loss which is prevalent in industrial indoor environments. (2) Effective population initialization, crossover and mutation, parallel computing as well as dedicated speedup measures are leveraged in the GA design for efficiently solve this TPC problem on a large scale. (3) The proposed TPC solution additionally includes powering-on/off mechanism. It does not require any modification on the wireless standard, which makes it cost-effective for large-scale industrial deployment. (4) Besides numerical experiments, four COTS Siemens industrial APs are deployed with a central control computer system for model and algorithm validation in a small-scale industrial environment.

The rest of this paper is organized as follows. Sect. 2 formulates this TPC problem in an integer programming model. Sect. 3 proposes the GATPC (genetic algorithm based TPC) algorithm to solve this model. Sect. 4 validates this model and GATPC in a small empty industrial environment. Sect. 5 performs numerical experiments in two vehicle manufacturers' indoor environments, standing for a typical small and large metal-dominated industrial indoor environment, respectively. Sect. 6 draws conclusions.

## 2. Problem formulation

The problem under investigation is optimal TPC of a dense WLAN in a metal-dominated industrial indoor environment. This IWLAN is over-dimensioned such that redundant APs exist for double full coverage. As a result, it is unnecessary for all APs to work at the maximal transmit (Tx) power level. There-fore, potential remains to minimize each AP's Tx power, including powering off.

A solution to this problem is denoted by $\vec{p}$. It is a vector of the Tx power levels of all over-dimensioned APs (denoted as $A$), including the decision of powering off certain APs.

### 2.1 Environment

A target rectangular environment is two-dimensional 2D, i.e., horizontal and vertical. It is represented by its two extreme 2D points: the upper left point (*xMin*, *yMin*) and the bottom right (*xMax*, *yMax*). It is discretized into $gs \times gs$ small grids, where *gs* is the grid size that is preset as an input of the model. Each grid point (GP) is represented by its upper-left point, and denoted as $gp_i$, where *i* is a unique index for each GP. A lexicographical order is applied to all the GPs:

$$(x0, y0) < (x1, y1) \Leftrightarrow x0 < x1 \vee \left( x0 = x1 \wedge y0 < y1 \right) \tag{1}$$

Consequently, a target environment is described by a set of ordered GPs denoted as $\Omega$. The GP index *i* within $\Omega$ starts from one, corresponding to the extreme point (*xMin*, *yMin*) of this environment. It increases one by one until reaching $|\Omega|$ following the lexicographical order. Then the set of GPs is denoted by their index $I = \left\{ 1, 2, ..., |\Omega| \right\}$. The following formula determines the size of $\Omega$:

$$|\Omega| = ceil \left[ (xMax - xMin) / gs \right] \times ceil \left[ (yMax - yMin) / gs \right] \tag{2}$$

A receiver (Rx) is placed on each GP except the ones where APs are placed. The received power in the downlink is considered to enable the calculation of an AP's coverage. For an Rx, different physical bitrate requirements have different requirements on the lowest received power, named threshold (*THLD*). The quantified relation can be found in [1].

The *i*-th GP is considered covered by the *j*-th AP, if an Rx on this GP connects to this AP and receives power values that are higher than or equal to the threshold during at least 99% of the time. This is formulated as follows:

$$\alpha_{ij} = \begin{cases} 1, \text{ if } P_{ij} \geq THLD \\ 0, \text{ otherwise} \end{cases}, \forall i \in I, \forall j \in J \tag{3}$$

where $\alpha_{ij}$ is the logical coverage variable for the *i*-th GP and *j*-th AP, and $P_{ij}$ is stable power (dBm) that an Rx on the *i*-th GP receives from the *j*-th AP at least 99% of the time. The coverage of an AP is represented by the GPs that are covered by this AP.

## 2.2 Over-Dimensioned Access Points

In total, $|J|$ APs are over-dimensioned with a minimal separation distance in the environment, where $J$ is the set of AP index which varies from one to the total number of APs ($|A|$ or $|\vec{p}|$), i.e., $J = \{1, 2, ..., |A|\}$.

In a TPC solution $\vec{p}$, the APs are regrouped into a set of APs that are powered off ($J_{off}$) and a set of APs that are powered on with a certain power value ($J_{on}$):

$$J = J_{on} \cup J_{off} \tag{4}$$

where the AP indices in $J_{on}$ and $J_{off}$ are still these in $J$.

All APs are of the same type. They have the same TPC range $P$ (in dBm) and step $\delta p$ (in dB), i.e., $P = \{P_{\min}, P_{\min} + \delta p, P_{\min} + 2\delta p, ..., P_{\max}\}$. In total, there are $N_P$ different Tx power values in $P$, except the possibility of powering off. Therefore, $P_j \in P$ if $j \in J_{on}$, where $P_j$ is the Tx power of the *j*-th AP. $P_j$ is not considered if the *j*-th AP is powered off (Table 1).

**Table 1**

Mapping between physical power, power state, and the digital transmit power level of an access point

| Transmit power set $P$ (dBM) | Power state | Transmit power level set $\hat{P}$ |
|:---:|:---:|:---:|
| - | Off | 0 |
| $P_{\min}$ | On | 1 |
| $P_{\min} + \delta p$ | On | 2 |
| $P_{\min} + 2\delta p$ | On | 3 |
| ... | On | ... |
| $P_{\max}$ | On | $N_P$ |

As indicated in Table 1, with the possibility of powering off, the TPC range $P$ is discretized into $\hat{P}$, which is a dimensionless set of all possible Tx power levels, i.e., $\{0, 1, 2, ..., N_P\}$. The discretized Tx power level of the $j$-th AP is denoted as $\hat{P}_j \left( \in \hat{P} \right)$. Specifically, the level zero stands for powering off. The level one represents the minimal Tx power level if an AP is powered on, and so on, until the maximal Tx power level $N_P$ if an AP is powered on. Consequently, the digital variable $\hat{P}_j \left( \forall j \in J \right)$ can be used to represent all the possible power states of an AP: powering off/on and if powering on, which Tx power this AP works at.

*2.3 Path Loss*

In [19], simple propagation scenarios are used, where the signal attenuation is essentially determined by the distance. In the proposed TPC model, a one-slope PL model considering metal obstacle shadowing loss along the propagation path is considered for accurate PL calculation. In total, there are $N_o (\geq 0)$ dominant metal obstacles in the investigated environment. This PL model is formulated as:

$$PL(d_{ij}) = PL0 + 10n \log_{10}(d_{ij}) + OL_{ij} + \xi \tag{5}$$

where $PL0$ (in dB) is the PL at the distance of one meter, $n$ is the PL exponent which is a dimensionless parameter indicating the increase of PL with the distance, $d_{ij}$ is the distance (in m) between the Rx placed on the $i$-th GP and the $j$-th AP, $OL_{ij}$ is the total obstacle loss (in dB) caused by the metal obstacles that block the line between the Rx placed on the $i$-th GP and the $j$-th AP, and $\xi$ (in dB) is the deviation between the measurement and the model, which is attributable to shadowing.

For an investigated environment, it assumes that the obstacle locations are fixed. The deviation $\xi$ in Eq. (5) follows a Gaussian distribution, with a mean of zero and a standard deviation $\sigma$. The gain and margin are considered in the link budget calculation to be more realistic, which is not taken into account in [19]. The total gain $G$ (in dB) is the sum of the AP transmitter's gain and the Rx's gain. The margin $M$ (in dB) is the sum of shadowing, fading and interference margin.

The total obstacle loss between the Rx on the $i$-th GP and the $j$-th AP is calculated in the following two equations. Eq. (6) iterates all the dominant obstacles in the environment and accumulates the additional PL caused by the obstacles that blocks in the line-of-sight (LoS) radio propagation from the $j$-th AP to the Rx on the $i$-th GP. Eq. (7) defines the logical signal blockage variable $\beta_{ij}^k$. If the $k$-th dominant obstacle has the shadowing effect on the LoS radio propagation from the $j$-th AP to the Rx on the $i$-th GP, it equals one. Otherwise, it equals zero. The calculations defined by Eqs. (6, 7) are only limited to APs that are powered on.

$$OL_{ij} = \sum_{k=1}^{N_o} \beta_{ij}^k \cdot OL_k, \ \forall i \in I, \ \forall j \in J_{on}, \ \forall k \in \{1, ..., N_o\} \tag{6}$$

$$\beta_{ij}^k = \begin{cases} 1, \text{if the } k\text{th metal blocks the line between} \\ \quad \text{the } i\text{th GP and the } j\text{th AP} \\ 0, \text{otherwise} \end{cases}, \forall i \in I, \ \forall j \in J_{on}, \ \forall k \in \{1, ..., N_o\} \tag{7}$$

Furthermore, compared to most coverage-related optimization problems that only rely on a 2D environment [1, 8-10, 22, 24], an obstacle is modeled as a 3D geometrical model in the decision making of LoS propagation between a GP-AP pair (i.e., the logical signal blockage variable $\beta_{ij}^k$). Both the $j$-th AP and the Rx placed on the $i$-th GP have their own heights. An obstacle has a 3D dimension of length × width × height. An obstacle blocks the LoS propagation as long as part of it crosses the straight line between the top of the $j$-th AP and the top of the Rx on the $i$-th GP. A detailed discussion on the 3D obstacle loss calculation can be found in [25].

If the $j$-th AP is powered on with the Tx power $P_j$, the maximal distance this AP can cover ($d_{j\max}$) can then be calculated, without considering the additional shadowing effects that may be caused dominant obstacles. For an AP that is powered off, $d_{j\max}$ is zero, indicating that it cannot cover any GP. This is formulated as follows:

$$d_{j\max} = \begin{cases} 10^{\left(\frac{P_j + G - M - THD - PL0}{10n}\right)}, \forall i \in I, \ \forall j \in J_{on}, \ P_j \in P \\ 0, \ \forall j \in J_{off} \end{cases} \tag{8}$$

*2.4 Interference*

An inevitable goal of Tx power management for a dense WLAN is the interference among APs or radio pollution. While dedicated frequency planning is out of scope in this paper, it is assumed that non-overlapping channels are effectively allocated to the dense APs. If an Rx on the $i$-th GP connects to the $j$-th AP ($j \in J_{on}$), the interference ($I_{ij}$, in dBm) to this Rx is then all the power this Rx can sense from the other APs that are powered on ($\forall j' \in J_{on}, j' \neq j$) [24, 26]. The interference calculated this way is also interpreted as noise [24]. If an AP is powered off, it is not considered by this calculation. This is formulated in the following two equations:

$$I_{ij} = 10 \cdot \log_{10} \sum_{j' \in J_{on}} 10^{P_{ij'}/10}, \ \forall i \in I, \ \forall j, j' \in J_{on}, j' \neq j, \ P_{j'} \in P \tag{9}$$

$$P_{ij} = P_j + G - M - PL(d_{ij}), \forall i \in I, \ \forall j \in J_{on} \tag{10}$$

The worst case is that all APs are powered on with the maximal Tx power ($P_j = P_{\max}, \forall j \in J_{on}, J_{on} = J$). Then the maximal interference ($I_{ij\max}$, in dBm) to an Rx can be calculated as follows:

$$I_{ij\max} = 10 \cdot \log_{10} \sum_{j' \in J_{on}} 10^{P_{j'}/10}, \forall i \in I, \forall j, j' \in J_{on} = J, j' \neq j, P_j = P_{\max} \tag{11}$$

*2.5 Transmit Power Control*

The TPC model is minimization of normalized total interference (Sect. 2.4) under the constraint of full coverage (Sect. 2.3) of a metal-dominated industrial environment (Sect. 2.1) which is deployed with over-dimensioned APs (Sect. 2.2). It is described in the following three formulae.

$$Objective: \min_{p_j} \left( \frac{\sum_{i=1}^{N_{GP}} \sum_{j=1}^{|A|} 10^{I_{ij} \cdot \gamma_{ij}/10}}{\sum_{i=1}^{N_{GP}} \sum_{j=1}^{|A|} 10^{I_{ij\max} \cdot \gamma_{ij}/10}} \cdot 100\% \right), \forall \widehat{p}_j \in \widehat{p}, \forall i \in I, \forall j \in J \tag{12}$$

*s. t.*:

$$\sum_{j=1}^{|A|} \alpha_{ij} \geq 1, \forall i \in \mu \cdot I \tag{13}$$

$$\gamma_{ij} = \begin{cases} 1, \text{if Rx on the } i\text{th GP connects to the } j\text{th AP} \\ 0, \text{otherwise} \end{cases}, \forall i \in I, \forall j \in J \tag{14}$$

Eq. (12) sets the object of TPC as minimizing the normalized interference (in mW) in the whole over-dimensioned network. The essential variable that is tunable for this optimization is the Tx power level of each AP ($\widehat{P}_j, \forall j \in J$) deployed in the environment.

Eq. (13) sets the constraint that a percentage $\mu$ of all the GPs must be covered by at least one AP, i.e., a coverage rate $\mu$ ($\mu \in (0,1]$) must be ensured in the target environment.

A logical variable of AP connection $\beta_{ij}$ is introduced in Eq. (14). If an Rx can sense multiple APs that are powered on, it connects to the one that achieves the highest received power at this Rx. If there are multiple APs that have the same highest received power at this Rx, the Rx randomly connects to one of these APs. An Rx can connect to at most one AP, while an AP can have multiple Rx that connect to it. While RSSI or received power of a client plays a vital role in handover and AP association [27], further discussion on client-AP association mechanism is out of scope in this paper.

Overall, the entire TPC model is mathematically formulated by Eqs. (1-14), and named the interference minimization based TPC model (IM-TPC). It is considered as large-scale if the target industrial indoor environment has a large size ($> 10000 \text{ m}^2$) and $gs$ is small (within several meters). Otherwise, it is considered as small-scale.

## 3. Solution algorithm

A GA is well known by giving a global optimal or near-optimal solution within a reasonable time period. It has been successfully applied to solve a number of industrial energy-related optimization problems. This is illustrated as the energy-cost-aware production scheduling based on GA optimization [27], where production jobs of a unit process are scheduled along the time span such that the total energy cost of these jobs is minimized under the volatile electricity price and no tardiness is produced compared to a due date.

The TPC problem investigated in this paper is intrinsically similar to the above scheduling problem. The APs' Tx power levels that remain to be tuned and the constraint of one full coverage layer correspond to the time frames that need to be scheduled to all jobs and the constraint of completing all jobs before the due date, respectively.

This GA based TPC algorithm is named GATPC. All the coverage-related calculations use the PL model in Eq. (6) that additionally considers the obstacle loss. Generally, covered GPs refer to a set of GPs that are covered by at least one AP. In contrast, uncovered GPs refer to a set of GPs that are not yet covered by any AP. Five definitions are further given as follows to facilitate the logical presentation of GATPC in the subsections.

**Definition 1**: *covered GPs* refer to all the GPs that are covered by at least one AP.

**Definition 2**: *new covered GPs* of a given AP refer to a set of GPs that are not yet covered, but can be covered by this AP at its current Tx power level.

**Definition 3**: *blank GPs* refer to a set of GPs that are not yet judged whether they can be covered by any AP at the current Tx power level.

**Definition 4**: a *GP-AP link* indicates that the investigated GP can be covered by the investigated AP at its maximal Tx power level. It thus shows an AP's potential to cover a GP.

**Definition 5**: the *nearest potential AP* of an uncovered GP is the AP that is the nearest to this GP among all the APs that have *GP-AP links* with this GP.

### 3.1 Parallel Genetic Algorithms

The speedup issue is sensitive to a large-scale optimization, including a GA search. A conventional GA structure can be found in [27]. The fundamental GA operations include: (1) initial population generation, where a fixed size of individual solutions are generated in a random manner; (2) crossover, which swaps part of genes of two chromosomes (i.e., individual solutions); (3) mutation, which swaps genes (i.e., part of a solution) of a chromosome; and (4) elitism, which remains a fixed size of the best individuals in a generation to the child generation.

All the aforementioned and fitness calculation of all individuals in a generation exhibit a common characteristic for applying "map-and-reduce" [28] or "divide-and-conquer" [22, 29, 30] parallel computation strategy: each operation contains multiple independent sub-operations of the same type and with different individuals. Therefore, the sub-operations can be conducted in parallel, such as by multithreads of a processor [31]. The results of sub-operations are then collected one by one at the end of each sub-operation. As a result, the GA search gains speedup since multithreads physically work in parallel in different cores of a processor.

However, a special attention should be paid to the parallelism of multiple crossover operations. Normally, two individuals should be selected from the entire population for one crossover operation, in order to ensure that the better individuals have higher probability to be involved in breeding the child generation. Therefore, all the parallelized crossover operations should simultaneously have full access to the entire generation, in spite of the parallelism. Furthermore, the qualification check and potential correction of two new individuals in each crossover operation can also be parallelized (Sect. 3.4).

Although the GA general structure is ready for use, the GA solution encoding, initial population generation, crossover and mutation must be specifically defined, to link the GA structure to the optimization problem that is investigated. Regarding solving the TPC model, all these operation definitions share the common purpose of simultaneously minimizing memory usage and runtime for a large-scale TPC model.

*3.2 Solution Encoding*

As introduced in Sect. 2, a TPC solution is $\vec{p}$, a vector containing $\left|\vec{p}\right|$ discretized AP Tx power levels, including powering off (Table 1). The index of a value in $\vec{p}$ corresponds to the index of the AP that is over-dimensioned in the environment. The list of over-dimensioned APs is sorted by applying the lexicographical order (Eq. (1)) to the GPs on which these APs are placed. Therefore, the $j$-th value in $\vec{p}$ corresponds to the Tx power level of the $j$-th AP.

As each scalar in $\vec{p}$ can be computationally represented by a 32-bit integer, little memory is needed even for encoding a large-scale TPC solution. For instance, 100 TPC solutions for 100 APs only takes up 390 kB, which is ignorable for a modem PC equipped with an 8 GB RAM (random access memory).

*3.3 Population Initialization*

It is not obliged to generate all qualified initial individuals, since unqualified individuals will be either eliminated by the populated evolution or improved by the crossover and mutation operations. However, any generation of unqualified individuals will produce computation redundancy to the GA search and thus reduce the optimization efficiency. Especially for large-scale optimization, the computation time to get an acceptable solution is quite sensitive to computation redundancy. Hence, the proposed initial population generation algorithm aims to produce 100% qualified initial individuals.

Algorithm 1 describes two steps to randomly generate a qualified individual for the initial population. At step 1, a TPC solution is randomly generated (line 1) without considering the coverage constraint defined by Eq. (13).

Step 2 are lines 2-22 in Algorithm 1. The GPs' coverage information is updated by setting APs with these random Tx power levels (lines 2-4). If the required coverage rate cannot be satisfied, this random TPC solution will be corrected (lines 5-22). The algorithm iterates the *blank GPs* and searches for the potential APs that can cover them, so that the *blank GPs* can be reduced as many as possible which results the increase in coverage rate. In the minor case where a GP is shadowed by an obstacle such that it cannot be covered by any AP, this GP is removed from *blankGPs*, and this iteration continues (lines 12-15). Meanwhile, this algorithm tries to lower the Tx power of all APs as much as possible, by choosing the *nearest potential AP* and setting its Tx power to the minimal level that can cover the target *blank GP* (lines 11-16). Once an AP changes its Tx power, the GP coverage information is updated (lines 17-21).

---

**Algorithm 1** Generation of a random TPC solution (RTPC)

**Input**: none

**Output**: a qualified random solution $\vec{p}$ ensuring one full coverage layer

1.  $\vec{p} \leftarrow |A|$ random numbers $\epsilon \, \hat{P}$ ;
2.  *coveredGPs* $\leftarrow$ *new covered GPs* of APs $\epsilon \, A$ with $\vec{p}$
3.  *blankGPs* $\leftarrow \Omega$ ;
4.  remove *coveredGPs* from *blankGPs*;
5.  **while** (*blankGPs* $\neq \emptyset$ && $|coveredGPs| < \mu \cdot \Omega$)
6.      **for** $j \leftarrow 1:|A|$
7.          **if** ($\vec{p}(j) < N_p$)
8.              set up *GP-AP links* between *blankGPs* and *AP(j)* $\epsilon \, A$ ;
9.          **end if**
10.     **end for**
11.     find *nearest potential AP* of a random *GP* $\epsilon$ *blankGPs*;
12.     **if** (*nearest potential AP* == $\emptyset$)
13.         remove $1^{st}$ *GP* from *blankGPs*;
14.         **continue**;
15.     **end if**
16.     assign *nearest potential AP* with the minimal Tx power level that can cover a random *GP* $\epsilon$ *blankGPs*;
17.     remove *new covered GPs* of *nearest potential AP* from *blankGPs*;
18.     *coveredGPs* $\leftarrow$ *new covered GPs* $\cup$ *coveredGPs*
19.     **if** (Tx power level of *nearest potential AP* == $N_p$)
20.         remove *GP-AP links* between *nearest potential AP* and all the related GPs in *blankGPs*;
21.     **end if**
22. **end while**

---

**Algorithm 2** Crossover for GATPC

---

**Input**: two selected parent TPC solutions
**Output**: two child TPC solutions

1. *xMin* ← max (minimal horizontal coordinates of all APs in indiv1 and indiv2);
2. *xMax* ← min (maximal horizontal coordinates of all APs in indiv1 and indiv2);
3. *xCrossover* ← a random coordinate ∈ [*xMin, xMax*);
4. chop graphically *indiv1* and *indiv2* into two parts along the same vertical line *xCrossover*, respectively
5. *childIndiv1* ← 1$^{st}$ part of *indiv1* + 2$^{nd}$ part of *indiv2*;
6. *childIndiv2* ← 1$^{st}$ part of *indiv2* + 2$^{nd}$ part of *indiv1*;
7. **for** *indiv* ∈ {*childIndiv1, childIndiv2*}
8.    // Lines 2-23 in Algorithm 1, where $\vec{p}$ is the transmit power vector indicated by *indiv* in Algorithm 2
9. **end for**

---

**Algorithm 3** Mutation for GATPC

---

**Input**: a child TPC solution output by crossover and selected for mutation
**Output**: a new child TPC solution

1. *selectedAPs* ← find from the input individual the AP(s) that has/have the highest Tx power level ($N_p$);
2. **if** ($\left| selectedAPs \right| > 1$)
3.    *selectedAP* ← a random AP in *selectedAPs*;
4. **end if**
5. new $\vec{p}$ ← power off *selectedAP*;
6. // Lines 2-23 in Algorithm 1, except that $\vec{p}$ in Algorithm 1 is the new $\vec{p}$

---

To produce the entire initial population, Algorithm 1 is iterated for a number of times that is equal to the population size. It also serves as a random TPC solution generation algorithm (named RTPC) for benchmarking.

*3.4 Crossover*

A crossover operation enables two parent solutions to breed two new child solutions by swapping the parents' genes. Three steps are defined in Algorithm 2 for GATPC's crossover.

Step 1 (lines 1-3, Algorithm 2) generates a random vertical line on the map of the rectangular environment. This line serves as a crossover point. The lower and upper bounds for generating the horizontal coordinate of this vertical line are defined such that the two areas split by this line have at least one AP, respectively. The purpose is to make sure that each crossover is effective without producing any child solution that is exactly the same as one of their parents.

Step 2 (lines 4-6 in Algorithm 2) swaps the APs of the two input TPC solutions around the randomly generated vertical line. Each AP keeps its current Tx power level. APs on the left side of this vertical line in the first parent solution are combined with these on the right side of this vertical line in the second parent solution. This produces the first child solution. The second child solution is obtained by jointing the rest part of the two parent solutions.

Step 3 (lines 7-9 in Algorithm 2) checks whether each child solution achieves full coverage. If the environment is not yet fully covered, the uncovered GPs will be addressed one by one with their *nearest potential APs*. The whole procedure follows lines 2-23 in Algorithm 1, except $\vec{p}$ in Algorithm 1 is one of the two child solutions instead of being randomly generated.

*3.5 Mutation*

A GA is known as ensuring a global optimum of an optimization problem. A mutation operation plays a vital role to this end. The mutation of GATPC is defined by Algorithm 3, comprising two steps. Step 1 (lines 1-5) powers off all APs that have the highest Tx power level ($N_P$). This aims to increase the diversity in the solution space and essentially avoid the GA search to approach toward a local optimum. A new individual is then created at the end of step 1.

Step 2 (line 6 in Algorithm 3) is the same as step 2 in Algorithm 1 (lines 2-23). It corrects the new individual produced by step 1 with the "best effort", if the environment cannot be fully covered at the end of step 1.

*3.6 Additional Speedup Measures*

As aforementioned, the design of GATPC in the former subsections follows the idea of decreasing computation time and memory, to enable large-scale optimization for a general industrial scale. The following measures are further taken to speed up the GATPC by reducing the computation redundancy.

An AP's maximal coverage distance ($d_{j\max}, \forall j \in J$) is extensively calculated by Algorithms 1-3. To speedup, an AP's $d_{j\max}$ is calculated by Eq. (8) before the actual start of a GA search. In total, $N_p$ different $d_{j\max}$ values are pre-calculated according to $N_p$ different AP Tx power values of an AP, and stored as a constant vector. All the $d_{j\max}$-related calculation during the GA search process will then simply look up to this vector, instead of repeating the PL calculation millions of times.

*GP-AP links* of all APs are very frequently set up or removed in Algorithms 1-3, which requires an extensive iteration of all possible GP-AP pairs ($|\Omega| \cdot |A|$ in the worst case). This certainly becomes a tedious and time-consuming operation for a large-scale environment that has more than 10,000 GPs as well as at least dozens or hundreds of APs. The corresponding speedup measure consists of the following four sequential steps. (1) For the *j*-th AP, search in the aforementioned $d_{j\max}$ vector for its maximal coverage distance corresponding to the current Tx power level. (2) Set up a $d_{j\max} \times d_{j\max}$ rectangular region that is centered at the *j*-th AP. (3) Iterate the GPs within the former rectangular area and set up *GP-AP links* of the *j*-th AP. (4) Iterate all APs and conduct steps (1-3) in each iteration. The obtained speedup is especially significant for a large-scale environment, since the area to set up *GP-AP links* is substantially reduced from the entire environment to the $d_{j\max} \times d_{j\max}$ small square.

Besides, Algorithms 1-3 frequently judge whether an obstacle shadows the signal between the *j*-th AP and an Rx on the *i*-th GP, and then calculate the accumulated obstacle loss, i.e., Eq. (6, 7). The speedup measure is inspired from the fact that, for a certain GA search, all the obstacles and APs are static in terms of quantity and location. Consequently, the signal blockage between the *i*-th GP and the *j*-th AP can be judged before the GA search, and the corresponding obstacle loss (including zero loss) can be pre-stored in a table. The GA search will then only need to enquire the pre-stored table of obstacle loss by inputting the indexes of GP and AP, instead of judging on-the-fly.

Last but not least, Algorithms 1-3 very frequently judge whether a GP is covered by an AP at its current Tx power level, i.e., Eq. (3). Thereby, the PL calculation considering the shadowing effects of dominant obstacles, i.e., Eqs. (5-7), should extensively be performed. As a speedup measure, Eq. (3) is implemented in two sequential steps for the *i*-th GP and the *j*-th AP. (1) Look up to the aforementioned $d_{j\max}$ vector for the corresponding $d_{j\max}$ of the *j*-th AP. (2) Set up a $d_{j\max} \times d_{j\max}$ square that is centered at the *j*-th AP. (3) Calculate the PL between the *j*-th AP and the *i*-th GP, without considering the shadowing effects. (4) Look up to the aforementioned obstacle loss table, and add the obstacle loss to the PL that is obtained in step (3), and get the final PL value. (5) Obtain $P_{ij}$ with the final PL value and judge whether it is above the preset sensitivity threshold.

## 4. Experiment validation

The TPC model and the GATPC algorithm were validated in a small open industrial environment (10 × 10 m) in the factory hall of an AGV manufacturer, in Flanders, Belgium.

### 4.1 Configurations

A measurement control system [1] accommodating the GATPC and four Siemens® industrial APs (Scalance W788-2 M12) with individual power supply (Fig. 1a) were used.

More specifically, an AP has two radio ports (Fig. 1a). One was configured for measurements at 2.4 GHz and the other was configured for remote control at 5 GHz. For an AP, 44 dB attenuation was added to each of the three ports of the measurement radio, to mimic a larger environment needing four APs for double full coverage. Individual power supply plus an extension power cable was applied to every AP, to enable deployment without the distance limitation. The remote AP control was realized by SSH (secure shell). The

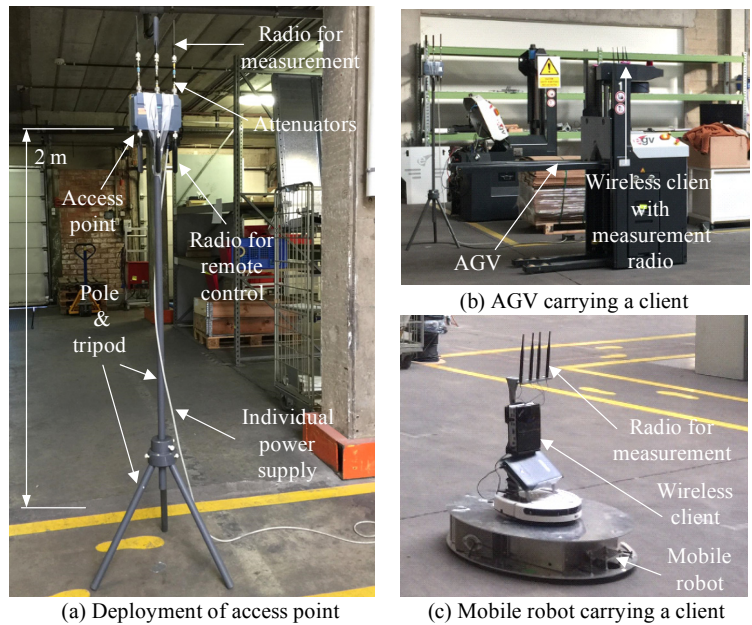

(a) Deployment of access point

(b) AGV carrying a client

(c) Mobile robot carrying a client

Fig. 1.  Experimental facilities, including a measurement control computer system, four commercial off-the-shelf industrial access points, an automated guided vehicle (AGV) with a wireless client, and a mobile robot with a wireless client.

**Table 2**

**Configurations of the measurement campaign**

| | |
|---|---|
| AP Tx power range with attenuation | -39:1:-27 dBm |
| WLAN standard | IEEE 802.11n |
| AP working frequency band | 2.4 GHz |
| AP remote control frequency band | 5 GHz |
| AP height | 2 m |
| AP1 location | (8, 40) m |
| AP2 location | (14, 31) m |
| AP3 location | (6, 35) m |
| AP4 location | (16, 36) m |
| Required physical bitrate of a wireless client | 24 Mbps |
| Required receiving sensitivity | -79 dBm |
| Mobility speed of the AGV and robot | 20 cm/s |
| Grid size for coverage monitoring | 1 m |
| Shadowing margin (95%) | 1 dB |
| Fading margin (99%) | 0 dB |
| Interference margin | 0 dB |
| GATPC stop criterion | 30 iterations |

central PC thus sent control wireless commands to an AP, such as setting the Tx power and powering on/off a radio.

The four APs were over-dimensioned on the boundary of the environment, such that each side was placed with one AP and double full coverage was planned [1]. The AP locations are indicated in Table 2, of which the coordinates are these used by the localization system of the AGV.

The coverage measurement facilities that were used have been introduced in [1] in detail. They mainly include a measurement control software system, two Zotac® mini-PCs as two individual wireless clients, four poles with tripods to support the APs at the height of 2 m (Fig. 1a), an AGV as a controllable mobile vehicle which carries one client on the top (Fig. 1b), a w-iLab.t mobile robot [32] which carries the other client on the top (Fig. 1c).

Instead of manual measurements, the two clients automatically kept on moving around in the environment and measuring the coverage of the AP that they connected to, and fed the collected samples back to the central PC for monitoring. These samples were stored in CouchDB database of the measurement control system. Samples from the same AP and within the same spatial grid were further aggregated to one value (dBm) to enable stable coverage monitoring (Sect. 2.1). For the minority of grids that might contain no sample, interpolation [1, 33] was applied based on the surrounding samples. Table 2 lists the key measurement configurations.

In total, 3745 RF power samples were collected. Regression [3] was applied to these data to build an empirical PL model formulated by Eq. (5), where $PL0$ was 39.87, $n$ was 1.78, and the obstacle loss $OL_{ij}$ ($\forall i \in I, \forall j \in J$) was zero dB due to the empty environment.. The R-squared value was 97.38%, indicating that the PL model was highly fitted to the samples.

*4.2 Validation results*

The TPC solution given by the GATPC algorithm is illustrated by Fig. 2a. AP1 and AP2 are powered on at -36 dBm and -27 dBm, respectively. AP3 and AP4 are both powered off. The colored GPs (grid points) represent the highest received RF power from the existing APs. All the received RF power values are above the required lowest sensitivity (-79 dBm, Table 2), indicating one full coverage layer in the environment. In a conventional full power-on scheme, all the four APs are simply powered on with the maximal Tx power (-27 dBm). In comparison, in the obtained TPC solution, AP1 decreases the Tx power to -36 dBm, and AP3 and AP4 are powered off.

The power states and Tx power levels of the four deployed APs were then set according to this optimal TPC solution. The coverage was monitored. As visualized in Fig. 2b, the received RF power values vary between -60.4 dBm and -78.8 dBm. They are above the threshold sensitivity (-79 dBm, Table 2), demonstrating that the environment is fully covered. Therefore, the solution given by GATPC are effective to satisfy the major constraint (i.e., coverage, Eq. (13)) of the TPC model.

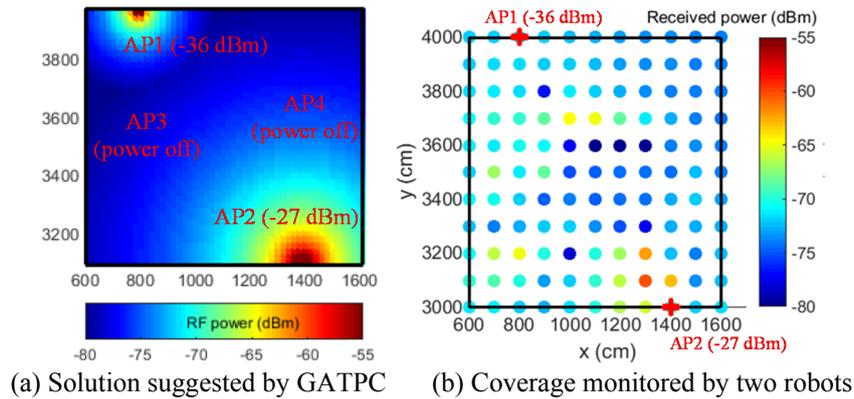

(a) Solution suggested by GATPC    (b) Coverage monitored by two robots

Fig. 2. Transmit power control solution given by the GATPC algorithm (Fig. 2a) and actual coverage monitored by an automated guided vehicle (AGV) and a mobile robot which carry wireless clients (Fig. 2b). The simulated and measured coverage maps are highly matched.

## 5. Numerical experiments

Numerical experiments were further conducted on the proposed GATPC algorithm. A 64-bit Win7 PC was used, with an Intel i5-3470 CPU (two 3.20 GHz single-thread cores) and an 8 GB RAM.

### 5.1 Configurations

The two investigated industrial indoor environments are a factory hall of an AGV manufacturer and a warehouse of a car manufacturer, both located in Flanders, Belgium.

The AGV factory hall (Fig. 3a) measures 102 m × 24 m. It represents a small-scale industrial indoor environment. It is placed with metal racks for component storage. AGVs of varying sizes are usually placed without moving and waiting for integration, maintenance, or shipment. Wide WiFi coverage is needed for AGV communication and Internet access of the workers' laptops.

The warehouse (Fig. 3b) measures 415 m × 200 m. It represents a large-scale industrial indoor environment. It is placed with metal racks at a height of nine meters. These racks are filled with wooden boxes that contains metal components. Wide WiFi coverage is required to support voice picking. Human pickers are equipped with microphones and earphones. They communicate with the control center via WLANs, to pick up from and place a stuff to a specific location.

For the TPC model, a metal rack in both cases is an obstacle that potentially causes evident shadowing effects to radio propagation. In the following numerical experiments, an obstacle measures 20 m × 3 m × 9 m. It can be placed either horizontally (i.e., the length side is parallel to the length side of the environment) or vertically (i.e., the length side is parallel to the width side of the environment). The direction and location of an obstacle are randomly generated by following a uniform distribution, while the entire part of a rack must be enclosed in the environment. The number of racks is an input of the TPC model. The GPs occupied by obstacles are not considered in the PL calculation.

The network parameters are summarized in Table 3, including the PL model, the AP transmitter, the receiver, and the environment. APs deployed by using the over-dimensioned algorithm such that tow full coverage layers are created in the target environment [1]. Each AP has 14 different Tx power levels, including powering off.

As pointed out in [22], the grid size ($gs$) influences the computational accuracy of coverage, $gs$ should be as small as possible without significantly compromising the computational complexity. Consequently, $gs$ is set as one meter, which is within 10 wave length (1.2 m). This means that the PL within this distance can be considered as constant without sacrificing the precision of PL calculation. The two parameters $PL0$ and $n$ of the one-slope PL model are same as these in Sect. 4. The PL caused by a metal rack (7.37 dB) is the mean of measured PL samples. The GA parameters are shown in Table 3.

Compared with the two environment sizes (68 m × 59 m and 12 m × 67 m) and around 30 APs involved in large-scale WLAN design in [24], our investigated two environments, especially the warehouse environment, show their hyper-large property for optimization.

*5.2 Effectiveness in Empty Environments*

The GATPC was first performed in the small-scale and large-scale environments without any presence of metal obstacles. Two other Tx power management schemes were used for benchmarking. One is the RTPC

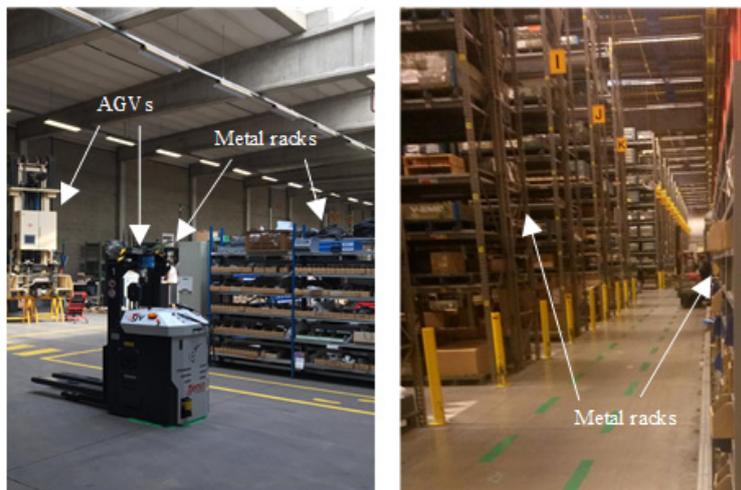

(a) Solution suggested by GATPC    (b) Coverage monitored by two robots

Fig. 3. The two industrial indoor environments for numerical experiments: a factory hall of an automated guided vehicle (AGV) and a warehouse of a car manufacturer. Both environments are placed with metal racks, which creates a challenge for radio propagation or robust wireless connection.

**Table 3**

**Configurations of the numerical experiments**

| Path Loss Model | |
|---|---|
| PL0 | 39.87 dB |
| $n$ | 1.78 |
| Shadowing margin (95%) | 7 dB |
| Fading margin (99%) | 5 dB |
| Interference margin | 0 dB |

| Access Point (AP) | |
|---|---|
| Height | 2 m |
| Gain | 3 dB |
| WiFi standard | IEEE 802.11n |
| Transmit power range | {-5:1:7} dBm |
| Locations | Output by over-dimensioning |
| Numbre of APs | 4 (small-scale environment) |
| | 75 (large-scale environment) |

| Wireless Client | |
|---|---|
| Height | 1.4 m |
| Gain | 2.15 dB |
| Required physical bitrate | 54 Mbps |
| Required minimal sensitivity | -68 dBm |

| Environment | | |
|---|---|---|
| Factory hall | Size (small scale) | 2448 m$^2$ (102 m × 24 m) |
| | Grid point number | 2600 |
| Warehouse | Size (large scale) | 83,000 m$^2$ (415 m × 200 m) |
| | Grid point number | 83,616 |
| Grid size ($gs$) | | 1 m |
| Radio frequency | | 2.4 GHz |
| Antenna type | | Omnidirectional |
| Metal rack size | | 20 m × 3 m × 9 m |
| Path loss caused by one metal rack | | 7.37 dB |

| GATPC algorithm | |
|---|---|
| Population size | 60 (small-scale environment), 100 (large-scale environment) |
| Elitism rate | 4% |
| Crossover rate | 70% |
| Mutation rate | 40% |
| Stop criterion | 50 iterations |

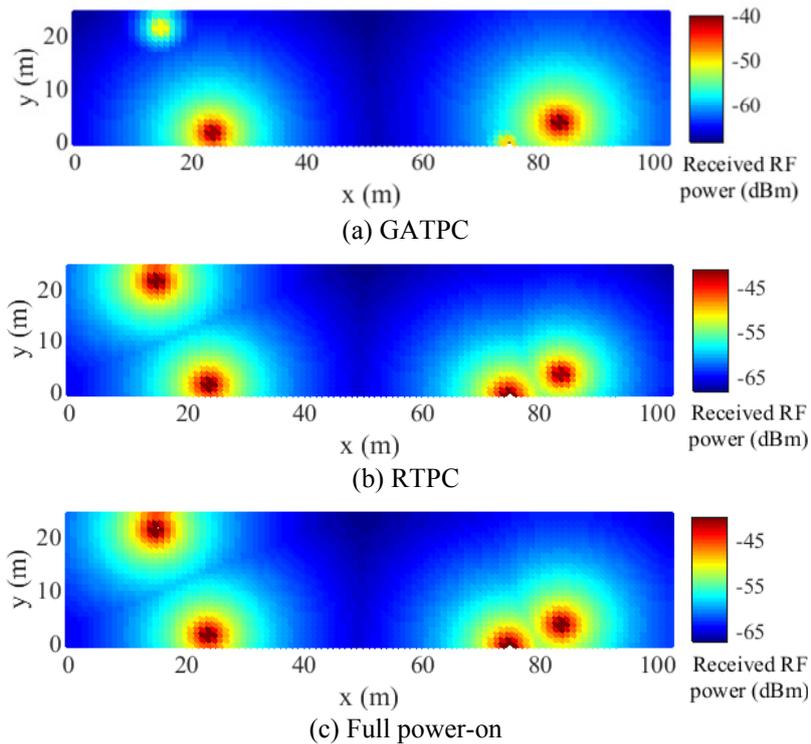

(a) GATPC

(b) RTPC

(c) Full power-on

Fig. 4. Three transmit power control schemes for an empty small-scale environment. The proposed GATPC algorithm is notably superior in reducing the transmit power or coverage of over-dimensioned wireless nodes while ensuring full coverage in the environment.

scheme (Algorithm 1). The other is the full power-on scheme, where all APs are powered on with maximal Tx power, i.e., no TPC is deployed.

### 5.2.1 Small-Scale Empty Environment

Overall, the GATPC is demonstrated to have notable superiority over the two benchmark schemes, in terms of reducing Tx power of wireless nodes and minimizing total interference in the network. More detailed results will be described as follows.

The GATPC significantly decreases the coverage of two APs while ensuring one full coverage layer in the environment. The Tx power of the four over-dimensioned APs is -4 dBm, 6 dBm, -3 dBm and 7 dBm, respectively, from the left to the right of Fig. 4a. In contrast, the RTPC exhibits very limited performance in reducing the redundant Tx power. Its coverage map (Fig. 4b) is close to that of the full power-on scheme (Fig. 4c). Its suggested Tx power is 6 dBm, 6 dBm, 6 dBm and 5 dBm, respectively.

As indicated in Table 4, the GATPC evidently reduces the total interference (-32.04 dBm). In comparison, the RTPC shows limited capacity in mitigating interference. Its interference level (-24.95 dBm) is close to that produced by the worst case (-23.83 dBm in full power-on scheme).

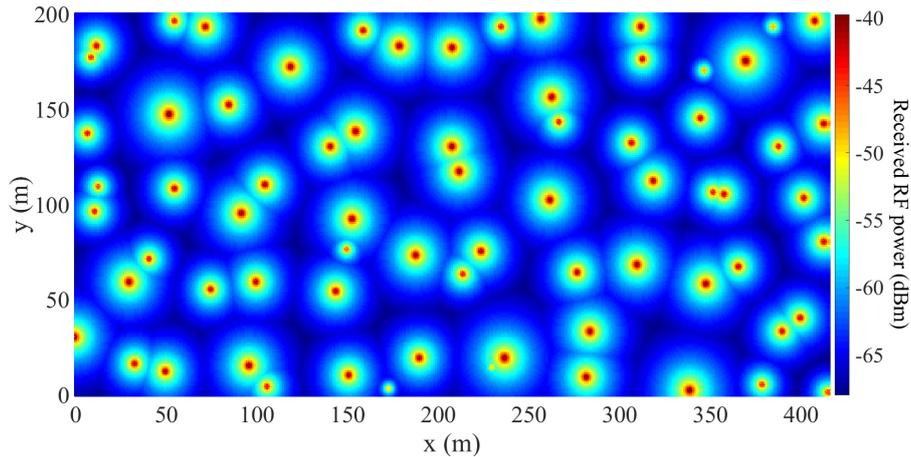

Fig. 5. Transmit power control solution suggested by the GATPC algorithm for an empty large-scale environment. Among the 75 over-dimensioned APs, 4 are powered off and most are set by transmit power lower than the maximum, while still having full coverage.

Besides, the runtime of GATPC is short (64 sec, Table 5). The RTPC has nearly zero runtime (Table 4), since optimization is not involved and the environment is small.

### 5.2.2 *Large-Scale Empty Environment*

The GATPC exhibits superior interference minimization performance in the large-scale empty environment. It achieves an interference level of -9.71 dBm in comparison to -9.29 dBm in the RTPC scheme and -7.02 dBm in the full power-on scheme (Table 4).

**Table 4 Interference of different transmit power control (TPC) schemes and runtime of RTPC**

| Environment type | | Small empty | Small obstructed | Large empty | Large obstructed |
|---|---|---|---|---|---|
| GATPC | Interference (dBm) | -32.04 | -26.59 | -9.71 | -10.02 |
| RTPC | Interference (dBm) | -24.95 | -24.97 | -9.29 | -9.56 |
| | Runtime (sec) | 0 | 0 | 103 | 131 |
| Full power-on | Interference (dBm) | -23.83 | -24.04 | -7.02 | -7.52 |

**Table 5 Speedup performance of the GATPC algorithm using high performance computing (HPC)**

| Runtime | Small empty | Small obstructed | Large empty | Large obstructed |
|---|---|---|---|---|
| With HPC (sec) | 64 | 73 | 102,841 | 167,504 |
| Without HPC (sec) | 94 | 172 | 3,866,700 | 4,670,300 |
| Reduction rate (%) | 31.9 | 57.6 | 97.3 | 96.4 |
| Speedup times | 0.5 | 1.4 | 37.6 | 27.9 |

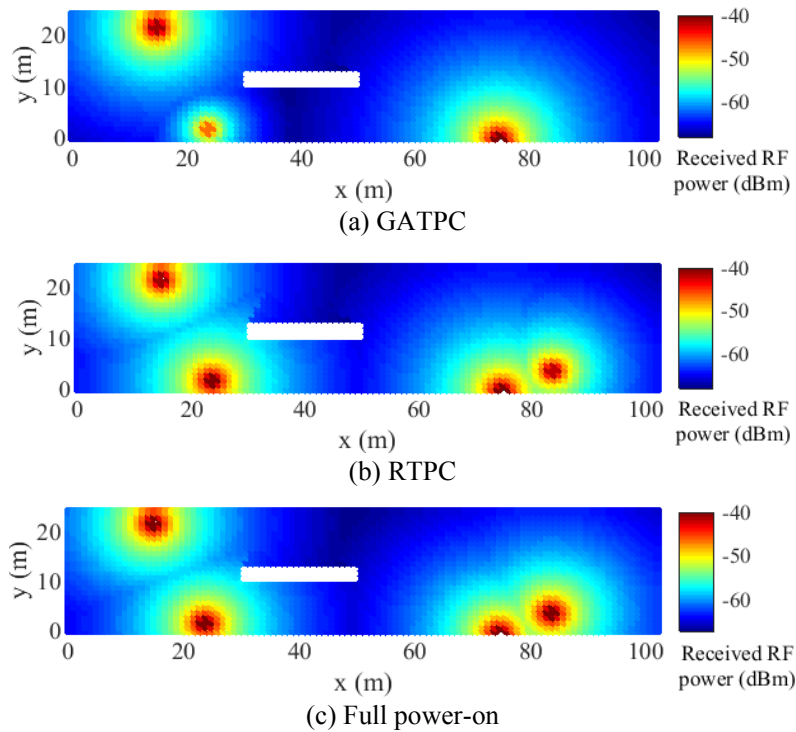

Fig. 6. Three transmit power control schemes for a small-scale environment placed with a metal rack. The proposed GATPC algorithm is evidently superior in reducing the transmit power or coverage of over-dimensioned wireless nodes while ensuring full coverage in the environment.

The GATPC is effective in AP Tx power reduction (Fig. 5). Besides the four powered-off APs, most of the powered-on APs are set by a Tx power level that is lower than the maximum (7 dBm, Table 3), and many are even set by a level which is very close or equal to the minimum (-5 dBm, Table 3).

The GATPC's runtime obviously increases (102,841 sec or about 28.5 h, Table 5) compared with that in a small-scale empty environment (64 sec, Table 5). This is explained by the 34 times larger area and the consequently 603 times more AP-GP pairs in the large-scale environment.

### 5.3 Effectiveness in Obstructed Environments

The GATPC was then performed in these small-scale and large-scale environments which are obstructed. To mimic the shadowing effects in industrial indoor environments, one metal rack (Table 3) was placed in the small-scale environment and ten in the large-scale environment with a 100% qualification rate. The two aforementioned benchmark schemes were also used to measure GATPC's performance.

### 5.3.1  Small-Scale Obstructed Environment

The GATPC obviously demonstrates superior TPC effectiveness in the small-scale obstructed environment. According to its output solution, it not only powers off one of the four APs, but also decreases the other two's Tx power (0 dBm and 6 dBm) while keeping the fourth one at the maximum (Fig. 6a). In contrast, the RTPC scheme exhibits little capacity to reduce the Tx power. Its output TPC solution (Fig. 6b) is quite close to that of the full power-on scheme (Fig. 6c).

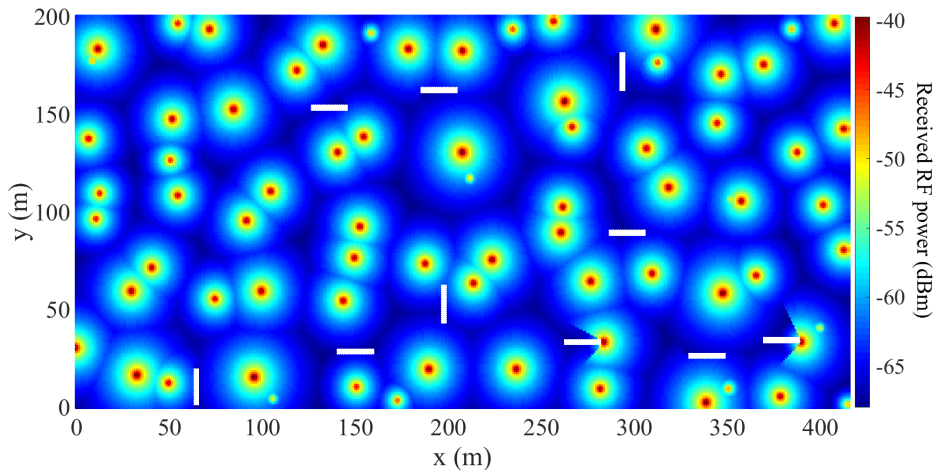

Fig. 7. Transmit power control solution suggested by the GATPC algorithm for an obstructed large-scale environment (the 10 while rectangles represent 10 randomly placed metal racks). Besides one AP that is powered off, many of the rest APs reduce their transmit power close to the minimum, while still ensuring full coverage.

The GATPC also shows up as the best in interference mitigation in the small-scale obstructed environment. It suppresses the inference down to -26.59 dBm (Table 4). This is lower than -24.97 dBm in the RTPC scheme and -24.04 dBm in the full power-on scheme (Table 4).

GATPC's runtime is short (73 sec, Table 5), due to the small scale of the investigated environment. It slightly increases compared to that in the small-scale empty environment. This is because of the additional obstacle loss calculation (Eqs. (6-7)), though the GPs taken up by the metal rack are excluded in the interference calculation. For the same two reasons (Sect. 5.2.1), the RTPC scheme has almost zero runtime.

### 5.3.2 *Large-Scale Obstructed Environment*

The GATPC also exhibits superiority in interference mitigation in the large-scale obstructed environment. It achieves a total interference level of -10.02 dBm, while the RTPC and full power-on schemes produce interference of -9.56 dBm and -7.52 dBm, respectively (Table 4).

The effectiveness of GATPC in TPC is further demonstrated in Fig. 7, which presents the corresponding coverage map. One AP is powered off and three APs are powered on with the minimal Tx power of -5 dBm. Among the APs that are powered on, many have Tx power levels that are lowered close to the minimum.

For the same two reasons explained in Sect. 5.2.2, the runtime of the GATPC rises to 167,504 sec compared to 73 sec in the small-scale obstructed environment (Table 5). Due to the additional obstacle loss calculation, it is also larger than that in the large-scale empty environment (102,841 sec, Table 5).

*5.4 Effectiveness in Speedup*

To further benchmark the GATPC's speedup performance, a derived version was used. It is the GATPC without high performance computing (HPC), including the parallel processing (Sect. 3.1) and speedup measures (Sect. 3.6). As it turned out to be very time-consuming to obtain an optimal solution in the large-scale environment (at the unit of months), the following means was taken to gauge its runtime.

First, the runtime to generate one random solution in the initial population was measured (at the scale of thousands sec). It was then multiplied by the population size to get the total runtime for population initialization. This derived GATPC version was rerun by enabling HPC in population initialization (Sect. 3.3) and followed by population evolution (Sects. 3.4 & 3.5) without HPC. Once the population went through one evolution and its corresponding runtime was got (at the scale of ten thousands sec), this algorithm stopped and this runtime was multiplied by the number of evolutions to obtain the runtime for evaluating the entire population. Finally, the estimated overall runtime was the sum of the runtime for the population initialization and that for the population evolutions.

The GATPC with HPC demonstrates significant speedup performance, as presented in Table 5. In the small-scale environment, its speedup times stay around one. The runtime of both algorithms are acceptable. However, in the large-scale environment, the speedup times boost to around 30. This makes it feasible to run the GATPC algorithm in a dramatically-reduced time horizon (1 - 2 days), in contrast to the infeasible runtime of the version with HPC. Given that a factory's major layout cannot change too frequently, this optimized runtime is acceptable from the perspective of adapting TPC to a factory layout while minimizing the network interference.

*5.5 Sensitivity of Qualification Rate*

As the "best effort" philosophy (Sects. 3.3-3.5) is applied in the GATPC algorithm, the qualification rate is investigated. This rate means the probability for this algorithm to intrinsically satisfy the TPC model's fundamental constraint (i.e., coverage, Eq. (13)) when all APs are powered on with the maximal Tx power. During this experiment, the correspondent interference was calculated each time when one metal rack was shifted to a different GP in the small-scale environment. This calculation was iterated over all the possible

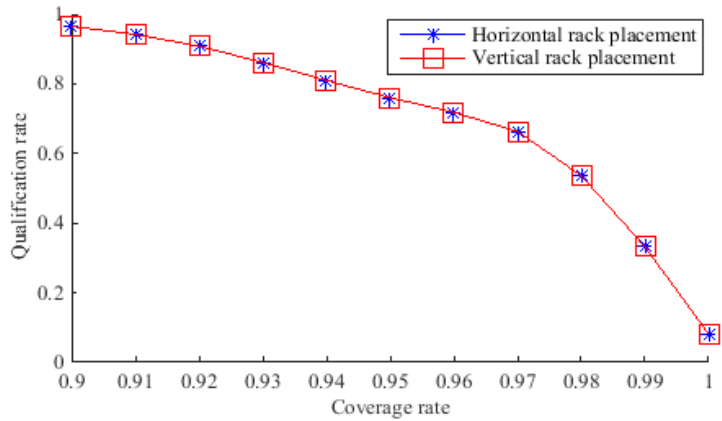

Fig. 8. Transmit power control (TPC) qualification rate to satisfy the required coverage rate in a metal-dominated environment. For each coverage rate, the rack location iterates over all the possible grid points with horizontal and vertical placement direction. As shown, 90% coverage can be guaranteed by the GATPC algorithm in more than 95% shadowing cases.

GPs. Consequently, the relationship between this qualification rate and the required coverage rate was captured.

As depicted in Fig. 8, 90% coverage can be guaranteed in more than 95% shadowing cases, demonstrating the GATPC's effectiveness in a general obstructed environment. The qualification rate is insensitive to the placement direction of a metal rack. It achieves as high as 96.5% at the coverage rate of 90%. It gradually decreases with the rising coverage rate, and finally drops to 8.1% in the case of full coverage. This decrease is explained by some specific rack locations on which a rack shadows all the potential over-dimensioned APs for some specific GPs. If a coverage level higher than 90% is desired for 95% shadowing cases, this improvement would rely on the over-dimensioning algorithm, instead of the TPC algorithm.

### 5.6 Sensitivity of Interference

The correlation between the interference and required coverage rate was further investigated under a varying number of metal racks placed in the small-scale environment. For each configuration, 30 independent runs were conducted and the average interference was collected, in order to get representative optimization results.

As indicated in Fig. 9, the interference declines from about -30 dBm at full coverage to -37 dBm at 50% coverage, regardless of the number of metal racks. This insensitivity to the number of metal racks implies that the limited number of GPs occupied by metal racks does not contribute much to the overall interference. This drop is explained by the continuously lower AP Tx power to satisfy the TPC model's coverage constraint which consistently becomes less strict. This is further demonstrated by Fig. 10, where the number of APs that are powered on generally increases with the decreasing coverage.

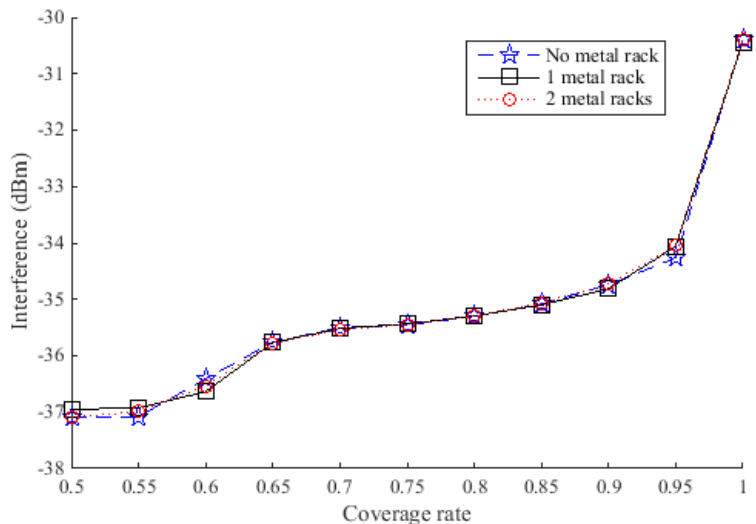

Fig. 9. Overall network interference under varying coverage rate and number of metal racks

Furthermore, the 10% coverage reduction from 100% to 90% contributes to more than 60% of the overall decreased interference (Fig. 9). This implies that lowering coverage requirement cannot be highly effective when the desired coverage rate drops below 90%. When the coverage falls in the range between 90% and 65%, the interference nearly remains stable. This is further proved by Fig. 10, where the number of APs that are powered off almost remains 1 when the coverage declines from 95% to 65%. In spite of the slight decrease in interference when the coverage continues to drop from 65% to 50%, the seriously-affected coverage should dramatically overweight this gentle decrease. Therefore, a coverage rate between 90% and 100% not only guarantees a high coverage level for wireless clients, but also is an effective range to control the overall interference.

## 6. Conclusion

With the ongoing trend toward Industry 4.0 or Industrial Internet in manufacturing, wireless technologies are penetrating into factories and warehouses, which include not only wireless sensors networks (WSNs) but also wireless local area networks (WLANs). This paper introduces a large-scale optimization problem of transmit power control (TPC) for dense industrial WLAN (IWLANs). It addresses the drawbacks of existing algorithms for coverage-related optimization problems, i.e., scalability, simplified coverage prediction model, incomplete power management schemes, and a lack of empirical validation.

To this end, this paper proposes genetic algorithm based TPC (GATPC). Instead of using the classical Boolean disk model, it integrates an empirical one-slope path loss (PL) model that considers three-dimensional shadowing effects in a metal-dominated industrial indoor environment. This PL model contributes to precise yet simple coverage prediction in the optimization algorithm. In GATPC, population initialization, crossover and mutation are designed to be effective such that GA search redundancy is reduced. High performance computing and dedicated speedup measures are used to further improve the efficiency of large-scale optimization.

The GATPC was experimentally validated in a small-scale industrial environment, and numerically demonstrated in both small-scale and large-scale industrial indoor environments. The solution quality of the GATPC was proven in terms of effectively conducting adaptive coverage and minimizing interference even in the presence of metal obstacles. The speedup performance of GATPC was measured to be as high as 37 times compared to the serial GATPC without speedup measures. The scalability of GATPC was demonstrated in the hyper-large optimization experiment compared to the state-of-the-art research.

The formulated TPC problem and the proposed GAPTC algorithm can also be applied to other types of wireless network besides WLANs, for instance optimal coverage maintenance of WSNs, which is one of the critical concerns in WSNs [9]. Regarding the future work, further speedup measures or high-performance algorithm design paradigms may be explored to additionally reduce the runtime of the GAPTC.

**Acknowledgement**

This research was supported by the ICON-FORWARD project with imec.